\documentclass[journal]{IEEEtran}
\ifCLASSINFOpdf
\else
\fi

\usepackage{amssymb}
\usepackage{amsmath}
\usepackage{makecell,rotating,multirow}
\usepackage{algorithm}
\usepackage{algorithmic}

\begin{document}
%
\title{Extreme Learning Machine with Local Connections}
%
%
%

\author{Feng~Li, Sibo~Yang, Huanhuan Huang, and Wei Wu
\thanks{W. Wu is the corresponding author (e-mail: wuweiw@dlut.edu.cn).}
\thanks{F. Li, S. Yang, H. Huang and W. Wu are with the School of Mathematical Sciences, Dalian University of Technology, Dalian 116024, China.}}

%
%

\markboth{}%
{Shell \MakeLowercase{\textit{et al.}}: Bare Demo of IEEEtran.cls for IEEE Journals}
%



\maketitle

\begin{abstract}
This paper is concerned with the sparsification  of  the input-hidden weights of ELM (Extreme Learning Machine). For ordinary feedforward neural networks, the sparsification  is usually done by introducing certain regularization technique into the learning process of the network. But this strategy  can not be applied for ELM, since the input-hidden weights of  ELM are supposed to be randomly chosen rather than to be learned. To this end, we propose a modified ELM, called ELM-LC (ELM with local connections), which is designed for the sparsification of the input-hidden weights as follows: The hidden nodes and the input nodes are divided respectively into several corresponding groups, and an input node group is fully connected with its corresponding hidden node group, but is not connected with any other hidden node group. As in the usual ELM, the hidden-input weights are randomly given, and the hidden-output weights  are obtained through a least square learning. In the numerical simulations on some benchmark problems, the new ELM-CL behaves better than the traditional ELM.
\end{abstract}


%
\IEEEpeerreviewmaketitle

\section{Introduction}
%
%
%
%
\IEEEPARstart{F}{eedforward} neural networks (FNNs) have been widely used in many fields due to their outstanding approximation capability \cite{Haykin,Zurada1}. The most popular  learning method  for FNNs is the back-propagation (BP) algorithm \cite{Rumelhart,LeCunY}, which is essentially composed of numerous gradient descent searching steps. One of the drawbacks of this gradient-based learning method is its slow convergence.

Extreme learning machine (ELM) \cite{HuangGB2006,HuangGB2011,HuangGB2016} was proposed to speed up the convergence by randomly choosing, rather than iteratively learning, the weights between the input and hidden layers. ELM  was shown to be highly efficient and easy to perform \cite{HuangGB2006,HuangGB2008}. To guarantees the convergence of ELM,  the number of the hidden nodes is required to be greater than or equal to the number of the training samples, which is usually quite large in practice. Although it is not a necessary condition for the convergence, but indeed the  the number of the hidden nodes should be quite large. At least, it should be greater than the number of the input nodes (=the dimension of the input data) due to the following intuitive argument: For instance, a data set in lower (resp. higher) dimensional space is easier (resp. harder) to be classified when it is randomly mapped into a higher (resp. lower) dimensional space. Therefore, there will be very many input-hidden weights when the number of the input nodes is big.

Regularization methods are often  used in the BP learning process to  remove the redundant weights \cite{Fan1,Fan2}. But this strategy  is not good to apply here for removing the redundant input-hidden weights of ELM since, as mentioned above, the input-hidden weights of ELM are supposed to be randomly chosen rather than iteratively learned.

For a related work on the local connection of ELM, we mention a local receptive fields based ELM (ELM-LRF) \cite{LRFELM}. ELM-LRF is designed for the data sets of images, which are  important examples  of high dimensional  data sets. Each elementary visual feature usually lies in different positions of an image datum. A remarkably successful neural network for dealing with such kind of data sets is the convolutional neural network (CNN) \cite{KrizhevskyA,KavukcuogluK,PfisterT,LiJ}. Convolutional hidden nodes are introduced in CNN for extracting and locating the  elementary visual features in different places of the the input image. CNN combines the technologies of local receptive fields \cite{LeCunY1998,EslamiSA,TurcsanyD}, visual cortex \cite{HubelDH}, shared weights and pooling etc.. Like the usual feedforward neural networks, CNN is also trained by the BP method, and  needs huge computational capability to tune the parameters.
To overcome this difficulty,  ELM-LRF was proposed recently by combining the network structure of CNN with the learning strategy of ELM.   The efficiency of ELM-LRF has been proved empirically in a few applications and generalizations \cite{BaiZ,LvQ,WangY,HuangJ,LiF}.

In this paper, we concern with another class of input data: They are of comparatively high dimension but, not like the image data etc., each component of the input data represents a specific attribute. We call them HDNI (High Dimensional Non-Image) data. When we are trying to build up a neural network to work on HDNI data, the technologies such as visual cortex, shared weights and pooling etc. are no longer applicable.  But the idea of local receptive fields (or local connections) remains useful.

To deal with HDNI data,  we propose an ELM with local connections (ELM-LC), of  which the input and hidden nodes are divided into corresponding groups, and are connected in a group-group manner.
Let us elaborate the network structure of ELM-LC by a simple example. Suppose that the input dimension is nine and that the output dimension is $m$. As illustrated in Fig. 1, we divide the nine input nodes into three groups such that each group contains three input nodes. (For simplicity, the bias nodes are ignored in the discussion here.) Each group of input nodes is fully connected with a corresponding hidden node group containing four hidden nodes, but is not connected with the other hidden node groups. Then, the hidden nodes are simply fully connected with the output node. Each pair of input-hidden groups works like the input and hidden layers of a small ELM network. A hidden node group should contains more nodes than its corresponding input node group. All these small ELM networks collaborate with each other through the hidden-output weights to get the final output. As in the usual ELM, the input-hidden weights  are randomly given, and the hidden-output weights are obtained through a least square learning.

\begin{figure}[!hbpt]
\vspace{-1em}
\centering
\includegraphics[width=3.7in]{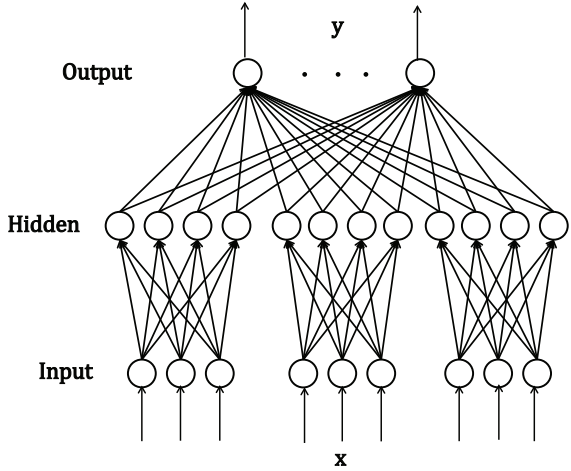}
\vspace{-1em}
\caption{Illustration of input-hidden local connections. }
\label{local}
\end{figure}

The idea of ELM-LC might be  illustrated intuitively by a famous Indian  fable ``The six blind men and the elephant": Each of the six blind men feels an elephant only by touching a separate  part of it, and draws a conclusion what the elephant is like. This fable tells us ``not to take a part for the whole". But on the other hand,  if the six gentlemen work together by  synthesizing their understandings for different parts of the  elephant, it is very likely for them to reach a complete picture of the elephant. The concept of ``local connection" in our method means that each blind man only touches a part, rather than the whole, of the elephant,  so as to   lighten his work load.  As a comparison, the task of ELM-LRF is to find and locate an elephant in a picture, where the elephant may appear in different places of the picture. And the task of ELM-LC is to identify an elephant by recognizing and synthesizing the different parts of it.

It is our expectation to get a sparse and robust network ELM-LC for HDNI data. Numerical simulations are carried out on some benchmark problems, showing that ELM-CL behaves better than the traditional ELM on HDNI data.

The remaining part of this paper is organized as follows. The description of the algorithm ELM-LC is presented in Section II. Supporting numerical simulations are provided in Section III. Some conclusions are drawn in Section IV.

\section{Description of the algorithm ELM-LC}
\subsection{A brief review of ELM}
ELM is a kind of single-hidden-layer feedforward neural network proposed by Huang et al. in \cite{HuangGB2006,HuangGB2011,HuangGB2016}. The key idea of ELM is to randomly generate the input-hidden weights and biases instead of tuning them iteratively, which speed up the learning process intensively and transform the original nonlinear problem into a linear problem.

\begin{figure}[!hbpt]
\vspace{-1em}
\centering
\includegraphics[width=3.7in]{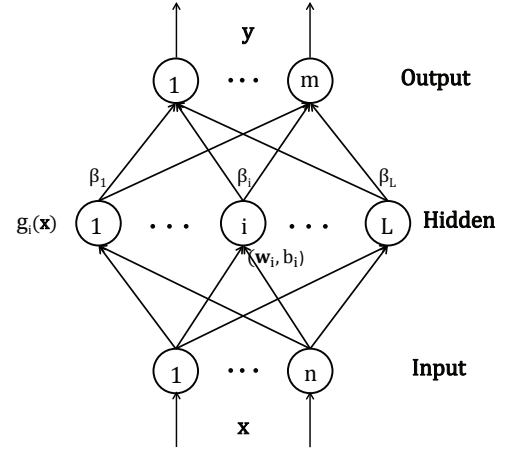}
\vspace{-1em}
\caption{Schematic diagram of fully connected ELM.}
\label{diagram}
\end{figure}

To elaborate, let us consider a feedforward neural network with $n$ input nodes, $L$ hidden nodes and $m$ output nodes as shown in Fig. \ref{diagram}. An input vector $\mathbf{x}\in \mathbf{R}^n$ is first converted to the hidden layer by a random  feature mapping. Then the output with respect to $\mathbf{x}$ is obtained by a linear mapping as follows:
\begin{equation}\label{EQ1}
\begin{split}
\mathbf{y}&=\sum_{i=1}^L\beta_i g_i(\mathbf{x})\\
          &=\sum_{i=1}^L\beta_i g(\mathbf{w}_i\cdot\mathbf{x}+b_i),
\end{split}
\end{equation}
where $\mathbf{y}\in \mathbf{R}^m$ is the output vector, $\beta_i=(\beta_{i1},\beta_{i2},\cdots,\beta_{im})^T\in \mathbf{R}^m$ is the outgoing weight vector from $i$-th hidden node, $\mathbf{w}_i=({w}_{1i},{w}_{2i},\cdots,{w}_{ni})^T\in \mathbf{R}^n$ is the ingoing weight vector of the $i$-th hidden node, $g(\cdot):R^1 \rightarrow R^1$ is a given activation function, and $b_i$ is the bias of the $i$-th hidden node.

Let $\big\{\mathbf{x}_j, {\mathbf{t}}_j\big\}_{j=1}^N\subset\mathbf{R}^{n}\times\mathbf{R}^m$ be a given sample set, where ${\mathbf{t}}_j$ is the ideal output for the input $\mathbf{x}_j$. The corresponding network outputs are
\begin{equation}\label{EQ2}
\mathbf{y}_j=\sum_{i=1}^L\beta_i g(\mathbf{w}_i\cdot\mathbf{x}_j+b_i), ~j=1,2,\cdots,N.
\end{equation}
The aim of ELM learning is to build up the network such that
$\mathbf{y}_j-{\mathbf{t}}_j=0$, or
 equivalent, to find $\mathbf{w}_i$, $b_i$ and $\beta_i$ such that
\begin{equation}\label{EQ4}
\sum_{i=1}^L\beta_i g(\mathbf{w}_i\cdot\mathbf{x}_j+b_i)={\mathbf{t}}_j, ~j=1,2,\cdots,N.
\end{equation}
The $N$ equations in (\ref{EQ4}) can be written compactly as:
\begin{equation}\label{EQ5}
\mathbf{H}\beta=\mathbf{T},
\end{equation}
Here,
\begin{equation}\label{EQ6}
\mathbf{H}=\begin{bmatrix}
g(\mathbf{w}_1\cdot\mathbf{x}_1+b_1) &\cdots & g(\mathbf{w}_L\cdot\mathbf{x}_1+b_L)\\
              \vdots                 &\cdots &                    \vdots\\
g(\mathbf{w}_1\cdot\mathbf{x}_N+b_1) &\cdots & g(\mathbf{w}_L\cdot\mathbf{x}_N+b_L)
\end{bmatrix}_{N\times L},
\end{equation}
\begin{equation}\label{EQ7}
\beta=\begin{bmatrix}
\beta_1^T  \\
 \vdots    \\
\beta_L^T
\end{bmatrix}_{L\times m},
\end{equation}
\begin{equation}\label{EQ7T}
\mathbf{T}=\begin{bmatrix}
\mathbf{t}_1^T  \\
 \vdots    \\
\mathbf{t}_N^T
\end{bmatrix}_{L\times m}.
\end{equation}

In ELM, the parameters $\mathbf{w}_i$ and $b_i$ ($i=1,2,\cdots,L$) are randomly generated rather than iteratively learned. Thus, the original nonlinear system is transformed into the linear system (4), which can be approximately solved by the usual least-square method:
\begin{equation}\label{EQ8}
\hat{\beta}=\mathbf{H}^\dag\mathbf{T},
\end{equation}
where $\mathbf{H}^\dag$ is the Moore-Penrose generalized inverse of matrix $\mathbf{H}$ \cite{Ben-IsraelA}:
\begin{equation}\label{EQ9}
\mathbf{H}^\dag=\left\{
\begin{array}{ll}
\mathbf{H}^T\mathbf{H}\mathbf{H}^T)^{-1}, & ~\text{if } N \leq L \\
\mathbf{H}^T\mathbf{H})^{-1}\mathbf{H}, & ~\text{if } N > L.
\end{array} \right.
\end{equation}

\subsection{ELM-LC (ELM with local connections)}
The proposed ELM with local connections (ELM-LC) is described below.
\begin{algorithm}[H]
\caption{: ELM-LC}
\begin{algorithmic}
\STATE {\bf Input.} {Input a given sample set $\big\{\mathbf{x}_j, {\mathbf{t}}_j\big\}_{j=1}^N\subset\mathbf{R}^{n}\times\mathbf{R}^m$, the hidden node number $L$, and  the group number $k$. $L$ should be large enough so as to satisfy the condition for the Grouping below.  }
\STATE {\bf Grouping.} Divide the $n$ input nodes roughly equally into $k$ groups, and similarly divide  the $L$ hidden nodes  into corresponding $k$ groups, such that  each hidden node group  contains at least one more node than its corresponding input group.
\STATE {\bf Random input-hidden weights.} Fully connect  each input node group with its  corresponding hidden node group by using randomly chosen weights, and the input node group has no connection with other hidden node group.
\STATE {\bf Computing of hidden-output weights.} Perform least square learning to compute the hidden-output weights as in the usual ELM.
\STATE {\bf Output.} Output the input-hidden and hidden-output weights.
\end{algorithmic}
\end{algorithm}

{\bf Remark}. We give a remark about the number of the input-hidden weights. For both ELM and ELM-LC, there are $n$ input nodes and $L$ hidden nodes. Obviously, the number of the input-hidden weights of ELM is $nL$. For ELM-LC, the input nodes and hidden nodes are divided into $k$ groups. When the nodes are equally divided, the number of the input-hidden weights of ELM-LC is $\frac{nL}{k}$.

\section{Numerical simulation}
Two artificial data sets and a UCI data set Facebook are used in our numerical simulation for regression problems. For classification problems, we use three UCI data sets:  Forest Types, Biodegradation,  and Ionosphere.
The usual Sigmoid function is used as the activation function for the hidden layer nodes. Our proposed ELM-LC is compared with the standard ELM. Each algorithm is run for ten trials for each data set.

To choose the number of hidden nodes for each data set, we perform ELM with different numbers of hidden nodes, and then choose the number of hidden nodes that achieves the smallest training error. The corresponding ELM-LC will then use the same number of hidden  nodes.

\subsection{Regression problem}
The following formula is used for generating the two artificial data sets:
\begin{equation}\label{functionvalue}
y=f(\mathbf{x})+\sigma\cdot \varepsilon,
\end{equation}
where $\mathbf{x}$ is a $d$-dimensional vector; $f(\mathbf{x})$ is an $R^d\rightarrow R^1$ function; and $\varepsilon$ is a standard Gaussian noise  added into the model to better test the generalization performance, which  is independent of $\mathbf{x}$ with noise level $\sigma=0.5$.  The two regression functions used for our simulations are defined respectively as follows.

\begin{equation*}
{\rm Function~I}:
\begin{array}{l}
\begin{split}
d&=12, \\
f(\mathbf{x})&=f(x_1,x_2,\cdots,x_{12})\\
&=\sum_{i=1}^{12}x_i\cdot \sin(x_i^2),
\end{split}
\end{array}
\end{equation*}
and
\begin{equation*}
{\rm Function~II}:
\begin{array}{l}
\begin{split}
d&=15, \\
f(\mathbf{x})&=f(x_1,x_2,\cdots,x_{15})\\
&=\sum_{i=1}^{15}x_i+ (-1)^i \ln(x_i^2).
\end{split}
\end{array}
\end{equation*}

For each function, 800 training samples are generated, where the input $\mathbf{x}$ of each sample is   uniformly distributed on $[-2,2]^d$, and its corresponding output $y$ is obtained by (\ref{functionvalue}).  200 test samples are generated similarly.

The experiments are also performed on a real world regression data set Facebook \cite{SinghK}. This data set is composed of 40949 training examples and 10044 test examples with 53 attributes and one target variable.

For Function I, the ELM has 12 input nodes and 32 hidden nodes, and the input and hidden nodes respectively are equally divided into 4 groups for ELM-LC. Similarly, the ELM for Function II has 15 input nodes and 30 hidden nodes, which are divided into 5 groups for ELM-LC. And the  ELM for the Facebook data set has 53 input nodes and 74 hidden nodes, which are divided into 9 groups for ELM-LC.

\begin{table*}[!t]
\renewcommand\arraystretch{1.3}
\caption{Comparison results of training error.}
\label{training1}
\scriptsize
\centering
\begin{tabular}{c|lll|lll|lll}
\hline
               &\multicolumn{3}{|c|}{Function I} & \multicolumn{3}{|c|}{Function II} & \multicolumn{3}{|c}{Fackbook}     \\
               &  Mean   &  Max    &  Min     &   Mean    &   Max     &    Min  &   Mean  &  Max    &    Min  \\
\hline
   ELM-LC       &  0.0423 & 0.0533  & 0.0283   &  1.5171   &	1.7466	  &  1.3100 & 0.0884  & 0.0898  & 0.0864   \\
   ELM        &  0.0892 & 0.1000  & 0.0835   &  2.1276   &	2.8241    &  1.8229 & 0.1059  & 0.1153  & 0.1005   \\
\hline
\end{tabular}
\end{table*}

\begin{table*}[!t]
\renewcommand\arraystretch{1.3}
\caption{Comparison results of Test error.}
\label{test1}
\scriptsize
\centering
\begin{tabular}{c|lll|lll|lll}
\hline
               &\multicolumn{3}{|c|}{Function I} & \multicolumn{3}{|c|}{Function II} & \multicolumn{3}{|c}{Facebook}     \\
               &  Mean   &  Max    &  Min     &   Mean    &   Max     &    Min  &   Mean  &  Max    &    Min  \\
\hline
   ELM-LC       &  0.0486 & 0.0652  & 0.0413   & 1.5568    & 1.8262    & 1.1891  & 0.0404  & 0.0619  & 0.0332   \\
   ELM        &  0.1017 & 0.1139  & 0.0871   & 2.2835    & 2.7076    & 1.8201  & 0.0786  & 0.1282  & 0.0425   \\
\hline
\end{tabular}
\end{table*}
The  training and test errors are shown in Tables \ref{training1} and \ref{test1}, where the mean, maximal and minimal values of the errors over the training or test samples are presented. We can  see that the  errors of ELM-LC are smaller than those of the standard ELM on the three data sets.

\begin{table}[!t]
\renewcommand\arraystretch{1.3}
\caption{Number of input-hidden weights of ELM and ELM-LC for regression problems.}
\label{weight1}
\scriptsize
\centering
\begin{tabular}{c|c|c|c}
\hline
               & {Function I}  &  {Function II} &  {Facebook}     \\
\hline
   ELM-LC       &  96           &   90           &  434           \\
   ELM        &  384          &   450          &  3922           \\
\hline
\end{tabular}
\end{table}

The Table \ref{weight1} shows the number of the input-hidden weights of ELM and ELM-LC. It can be seen that the number of the input-hidden weights of ELM-LC is much smaller than that of ELM.

\subsection{Classification problem}

The first data set for classification problem is the Forest Types data set \cite{JohnsonB}. Each datum has 27 attributes indicating certain  characteristics of the forest types. The data are divided into four classes: `s' (`Sugi' forest), `h' (`Hinoki' forest), `d' (`Mixed deciduous' forest) and `o' (`Other' non-forest land). The second data set is the Biodegradation data set \cite{Mansouri} with 41 molecular descriptors and two classes: ready biodegradable (RB) and not ready biodegradable (NRB). The third one is the Ionosphere data set \cite{Lichman} with 34 attributes and two classes: `Good' returns and `Bad' returns. More detailed information of these data sets is given in Table \ref{data}.

For the Forest Types data, the ELM has 27 input nodes and 36 hidden nodes, and these input and hidden nodes respectively are equally divided into 9 groups for ELM-LC. Similarly, for the Biodegradation data, there are 41 input nodes and 101 hidden nodes, which are roughly equally divided into 10 groups; And the Ionosphere data has 34 input nodes and 51 hidden nodes, which are equally divided into 17 groups.
\begin{table}[!t]
\small
\caption{Description of the data sets}
\label{data}
\centering
\begin{tabular}{ccccc}
\hline\noalign{\smallskip}
\multirow{2}{*}{Name} & \multirow{2}{*}{Features} & \multirow{2}{*}{Classes} & \multicolumn{2}{c}{Size}     \\
                &          &              &   Training     &  Test   \\
\noalign{\smallskip}\hline\noalign{\smallskip}
Forest Types    &  27   & d, h, s, and o  & 325    & 198    \\
Biodegradation  &  41   & RB and NRB      & 837    & 218    \\
Ionosphere      &  34   &  Good and Bad   & 250    & 101      \\
\noalign{\smallskip}\hline
\end{tabular}
\end{table}

\begin{table*}[!t]
\renewcommand\arraystretch{1.3}
\caption{Comparison of training accuracy}
\label{training2}
\scriptsize
\centering
\begin{tabular}{c|lll|lll|lll}
\hline
               &\multicolumn{3}{|c|}{Forest Types} & \multicolumn{3}{|c|}{Biodegradation} & \multicolumn{3}{|c}{Ionosphere}     \\
               &  Mean   &  Max    &  Min     &   Mean    &   Max     &    Min  &   Mean  &  Max    &    Min  \\
\hline
   ELM-LC       &  90.18  & 91.38   &  88.92   &  89.00    & 89.96     &  88.05  & 92.75   & 94.00   & 91.60   \\
   ELM        &  86.06  & 88.62   &  82.77   &  87.96    & 89.13     &  87.10  & 90.08   & 92.00   &  88.00   \\
\hline
\end{tabular}
\end{table*}

\begin{table*}[!t]
\renewcommand\arraystretch{1.3}
\caption{Comparison of test accuracy}
\label{test2}
\scriptsize
\centering
\begin{tabular}{c|lll|lll|lll}
\hline
               &\multicolumn{3}{|c|}{Forest Types} & \multicolumn{3}{|c|}{Biodegradation} & \multicolumn{3}{|c}{Ionosphere}     \\
               &  Mean   &  Max    &  Min     &   Mean    &   Max     &    Min  &   Mean  &  Max    &    Min  \\
\hline
  ELM-LC       &  90.55  & 91.92   & 89.39    &  83.76    & 85.32     & 82.11   & 97.03   & 99.01   & 96.04   \\
  ELM        &  85.56  & 88.89   &  82.83   &  82.47    & 83.49     &  80.28  & 96.73   & 98.02   &  95.05  \\
\hline
\end{tabular}
\end{table*}

The results of training and test accuracies are shown in  Tables \ref{training2} and \ref{test2}. We can see that ELM-LC achieves better training and test accuracies than  the usual ELM on the three data sets. From Table \ref{weight2}, it can be seen that the number of the input-hidden weights of ELM-LC is much smaller than that of ELM.

\begin{table*}[!t]
\renewcommand\arraystretch{1.3}
\caption{Number of input-hidden weights of ELM and ELM-LC for classification problems.}
\label{weight2}
\scriptsize
\centering
\begin{tabular}{c|c|c|c}
\hline
               & {Forest Types}  &  {Biodegradation} &  {Ionosphere}     \\
\hline
   ELM-LC      &  108          &   415          &  102           \\
   ELM         &  972          &   4141         &  1734        \\
\hline
\end{tabular}
\end{table*}

\section{Conclusion}
In this paper, we propose an ELM with local connections (ELM-LC). Its input and hidden nodes are divided into corresponding groups, and each input node group is fully connected with its corresponding hidden node group but is not connected with other hidden node groups. Hence, ELM-LC has sparse input-hidden weights compared with the usual fully connected ELM.
As in the usual fully connected ELM, the input-hidden weights are randomly chosen rather than iteratively learned, and the hidden-output  weights are obtained through a least square learning.
ELM-LC is designed for the so called HDNI data, which has comparatively high dimensional input data but, not like the image data etc., each component of the input data represents a specific attribute. Numerical simulations on two artificial data sets and four real world  UCI data sets show that ELM-LC achieves better learning and test (generalization) accuracies than the usual fully connected ELM with the same number of hidden nodes.

\section*{Acknowledgment}
This research was supported by the National Science Foundation of China (NO: 61473059, 61403056) and the Fundamental Research Funds for the Central Universities of China.

\end{document}